\documentclass[a4paper,twoside]{article}

\usepackage{epsfig}
\usepackage{subcaption}
\usepackage{calc}
\usepackage{amssymb}
\usepackage{amstext}
\usepackage{amsmath}
\usepackage{amsthm}
\usepackage{multicol}
\usepackage{pslatex}
\usepackage{apalike}
\usepackage{algorithm2e}
\usepackage[bottom]{footmisc}
\usepackage{hyperref} 
\usepackage{SCITEPRESS}     % Please add other packages that you may need BEFORE the SCITEPRESS.sty package.
\usepackage{tikz}
\usepackage{tabularx}
\usepackage{booktabs}
\usepackage{multirow}
\usepackage{graphicx}
\usepackage{soul}
\begin{document}

\title{Enhancing Visual Perception in Foggy Conditions via Multiclass Fog Density Modeling}

\author{\authorname{Mohamad Mofeed Chaar\sup{1}\orcidAuthor{0000-0001-9637-5832} and Galia Weidl\sup{1}\orcidAuthor{0000-0002-6934-6347}}
\affiliation{Connected urban mobility, Faculty of Engineering,\\ University of Applied Sciences, Aschaffenburg, Germany}
\email{\{MohamadMofeed.Chaar, Galia.Weidl\}@th-ab.de}
}


\abstract{Autonomous driving (AD) systems have advanced rapidly over the past decade; however, robust perception under adverse weather conditions remains a major challenge, particularly in dense fog. In this work, we investigate fog-aware perception using synthetically generated fog data derived from the Waymo dataset. To support fog simulation, depth images are generated using an iterative learning approach. We consider five fog-density levels: clear, light fog, moderate fog, heavy fog, and very heavy fog. Instead of training a single unified model across all conditions, we train separate perception models for each fog-density level. Experimental results show that density-specific training improves performance in severe fog conditions. In particular, for the very heavy fog class, recall improves from 0.076 to 0.232, corresponding to an absolute gain of 15.6 percentage points. These findings suggest that deploying multiple specialized models—rather than a single general-purpose model—can improve perception robustness for autonomous vehicles under challenging visibility conditions. Future work will extend this strategy to additional sensing modalities, including LiDAR and radar, and evaluate generalization across diverse weather scenarios.}

\onecolumn \maketitle \normalsize \setcounter{footnote}{0} \vfill

\section{\uppercase{Introduction}}
\label{sec:introduction}

Autonomous driving (AD) systems rely primarily on perception modules to interpret the surrounding environment and to enable downstream tasks such as object detection and tracking~\cite{Ref001}. Despite recent progress, perception performance remains highly sensitive to adverse weather conditions, particularly fog, which reduces scene visibility and significantly degrades image contrast. This limitation poses a critical challenge for robust AD deployment in real-world environments.

In addition, as the best of our knowledge, publicly available datasets provide limited or no fog-labeled data~\cite{Ref002}. Consequently, training and evaluating perception models under fog conditions remains difficult~\cite{Ref003}, and the lack of standardized fog annotations prevents systematic benchmarking. To address this gap, we propose synthesizing fog~\cite{Ref007} on the Waymo dataset~\cite{Ref004} and evaluating object detection~\cite{Ref005,Ref006} performance under multiple fog severity levels. Our approach is based on depth-aware fog rendering, where the fog intensity~\cite{Ref008} is computed as a function of the distance between the camera~\cite{Ref009} and the scene content.

A key practical challenge is that the Waymo dataset does not provide ground-truth depth images~\cite{Ref010}. Therefore, we generate depth maps using an Iterative Learning Approach (ILA)~\cite{Ref010}, enabling depth estimation from RGB images. These estimated depth maps are then used to synthesize fog consistently across frames and scenes. In this work, we develop a new fog generation formulation by extending the Koschmieder model through a matrix-based representation. This extension enables more flexible fog synthesis while preserving the physical intuition of atmospheric attenuation. For reproducibility, we provide an open-source implementation of our fog synthesis pipeline on GitHub\footnote{\scriptsize\url{https://github.com/Mofeed-Chaar/A-Mathematical-Extension-of-the-Standard-Equation-for-Synthe--sized-Fog-Rendering.git}}.

During fog synthesis, the fog density is controlled using the attenuation coefficient $\beta$, which allows generating multiple fog severity levels on the same dataset. Table~\ref{tab:tab1} summarizes the $\beta$ values used in our experiments on the Waymo dataset. 
\begin{table}[ht] 
    \caption{The corrisponding $\beta$ with fog density which operated on Waymo Datasets} \label{tab:tab1} 
    \centering 
    \begin{tabular}{|c|c|} 
    \hline 
    Fog Density & $\beta$ \\ 
    \hline 
    Clear & -- \\ 
    \hline 
    Light fog & 0.03 \\ 
    \hline 
    Moderate Fog & 0.06 \\ 
    \hline 
    Heavy fog & 0.09 \\ 
    \hline 
    Very heavy fog & 0.12 \\ 
    \hline 
    \end{tabular} 
\end{table}
This controlled fog generation enables training and evaluation of object detection models under different fog densities separately. As demonstrated in our experimental results, the proposed pipeline improves the recall performance by approximately 15\% compared to the baseline.

\section{\uppercase{Related Work}}\label{sec:Related Work}
Autonomous driving (AD) perception has been significantly improved through the use of various sensing modalities, including LiDAR~\cite{Ref011}, stereo cameras~\cite{Ref012,Ref013}, radar~\cite{Ref014}, and monocular camera systems. However, the performance of most optical sensors degrades severely under foggy conditions due to light scattering and reduced visibility. Radar is generally less affected by fog, but it provides limited spatial resolution and remains constrained in detecting small or distant objects~\cite{Ref015}.

To address these limitations, several works have investigated multi-sensor fusion strategies to enhance perception robustness. For example, the authors in~\cite{Ref016} integrated four complementary sensors, including a standard camera, LiDAR, gated camera, and radar, and synchronized their outputs to improve AD perception under severe weather conditions. Their approach demonstrated notable improvements, achieving a perception performance of 76.69 in heavy fog using a unified detection model.

In parallel, deep learning-based methods have become increasingly prominent for adverse-weather perception enhancement. TransWeather~\cite{Ref017} proposes a unified transformer-based image restoration framework capable of handling multiple weather degradations, such as rain, fog, and snow, within a single encoder--decoder architecture. By introducing intra-patch transformer blocks to capture fine-grained degradation patterns and learnable weather-type embeddings for adaptive restoration, TransWeather achieves state-of-the-art performance in terms of PSNR and SSIM while maintaining efficient model complexity.

Similarly, the work in~\cite{Ref018} introduces ECL-YOLOv11, a lightweight enhancement of the YOLOv11 detector designed to maintain detection accuracy under adverse weather while preserving real-time performance. The method incorporates edge-enhancement convolutions to preserve object contours, a context-guided multi-scale fusion neck (AENet) to improve detection of low-visibility targets, and a lightweight shared detection head (LDHead) to reduce computational overhead. Experimental results report improved mAP and precision compared to the YOLOv11 baseline, while sustaining very high throughput (approximately 230 FPS), making it suitable for real-time automotive applications.

Despite these advances, relatively few studies explicitly model fog conditions according to varying density levels. In our previous work, supported by CARLA-based simulation datasets~\cite{Ref019,Ref020,Ref021}, we proposed a fog-density-aware perception pipeline~\cite{Ref010}. Instead of treating fog as a binary condition, the approach discretizes fog into multiple density levels (0\%, 25\%, 50\%, 75\%, and 100\%) and trains separate object detection models for each level using automatically labeled synthetic data. The results showed that density-specific training improves precision and recall across object categories, while performance drops significantly when models are evaluated under mismatched fog conditions. This highlights the importance of condition-specific perception strategies for reliable operation in foggy environments.

Building upon this foundation, the present work extends fog-density-aware perception to real-world driving data by synthesizing realistic fog effects on real images. This reduces the reliance on purely simulated environments and mitigates the risk of overfitting to simulation artifacts, thereby improving the generalization and robustness of the proposed approach for real-world autonomous driving scenarios.

\section{\uppercase{Methodology}}\label{sec:Methodology}
\subsection{Depth Map}\label{subsec:Depth Map}

A depth map~\cite{Ref022} represents the distance from the camera to the scene points, where each pixel encodes the estimated depth value. In stereo vision systems, depth is computed using two calibrated cameras separated by a fixed baseline distance $B$. 

The depth information is obtained by measuring the pixel disparity, which corresponds to the horizontal displacement between matching points in the left and right images. Given the camera focal length $f$, the depth $Z$ of a pixel can be calculated as~\cite{Ref012}:

\begin{equation}\label{equ:01}
    Z = \frac{Bf}{d},
\end{equation}

where $d$ denotes the disparity value. This relationship indicates that depth is inversely proportional to disparity: pixels with larger disparity correspond to closer objects, while smaller disparity values indicate farther regions in the scene.

\subsection{Fog Synthesizing}
Due to the limited availability of open-source datasets annotated with explicit fog density levels, foggy images were synthetically generated using a physically motivated fog synthesis model based on an extension of the Koschmieder law~\cite{Ref023}. This model simulates atmospheric scattering by combining scene radiance attenuation with airlight contribution, and is formulated as:

\begin{equation}
I_{\text{fog}}(x) = I_{\text{orig}}(x) \cdot e^{-\beta D(x)} + A \cdot \left(1 - e^{-\beta D(x)}\right),
\label{eq:01}
\end{equation}

where:
\begin{itemize}
  \item \( I_{\text{fog}}(x) \) denotes the intensity of the synthesized foggy image at pixel \( x \),
  \item \( I_{\text{orig}}(x) \) denotes the intensity of the corresponding clear image,
  \item \( \beta \) represents the atmospheric scattering coefficient controlling fog density,
  \item \( D(x) \) is the depth (distance from the camera) at pixel \( x \),
  \item \( A \) is the global atmospheric light, commonly set to a constant value (e.g., \( A = [255, 255, 255] \)) to simulate white fog.
\end{itemize}

The fog density is directly controlled by the scattering coefficient \( \beta \): higher values of \( \beta \) correspond to denser fog conditions, leading to increased attenuation of scene radiance and stronger airlight effects.

\subsection{Iterative Learning Approach (ILA)}

The Iterative Learning Approach (ILA)~\cite{Ref012} is designed to address the problem of missing or invalid depth values by progressively refining depth supervision through multiple training iterations. The core idea of ILA is to use the predictions of a depth estimation model to iteratively improve the quality and completeness of the training targets.

Initially, a depth prediction model is trained using available RGB images and their corresponding depth maps, where missing depth values are typically encoded as invalid or zero pixels. After this first training stage, the model is employed to predict depth maps for the same RGB inputs. The predicted depth values are then used to fill the missing regions in the original depth maps, producing an enhanced depth dataset with reduced invalid areas.

This refined depth dataset is subsequently used as ground truth for retraining the model in the next iteration. The process of prediction, depth completion, and retraining is repeated for a predefined number of iterations. With each iteration, the amount of missing depth information decreases, allowing the model to learn from increasingly complete and consistent supervision.

\begin{figure*}[ht]
    \centering
    \begin{subfigure}{.5\textwidth}
    \centering
    \includegraphics[width=.95\linewidth]{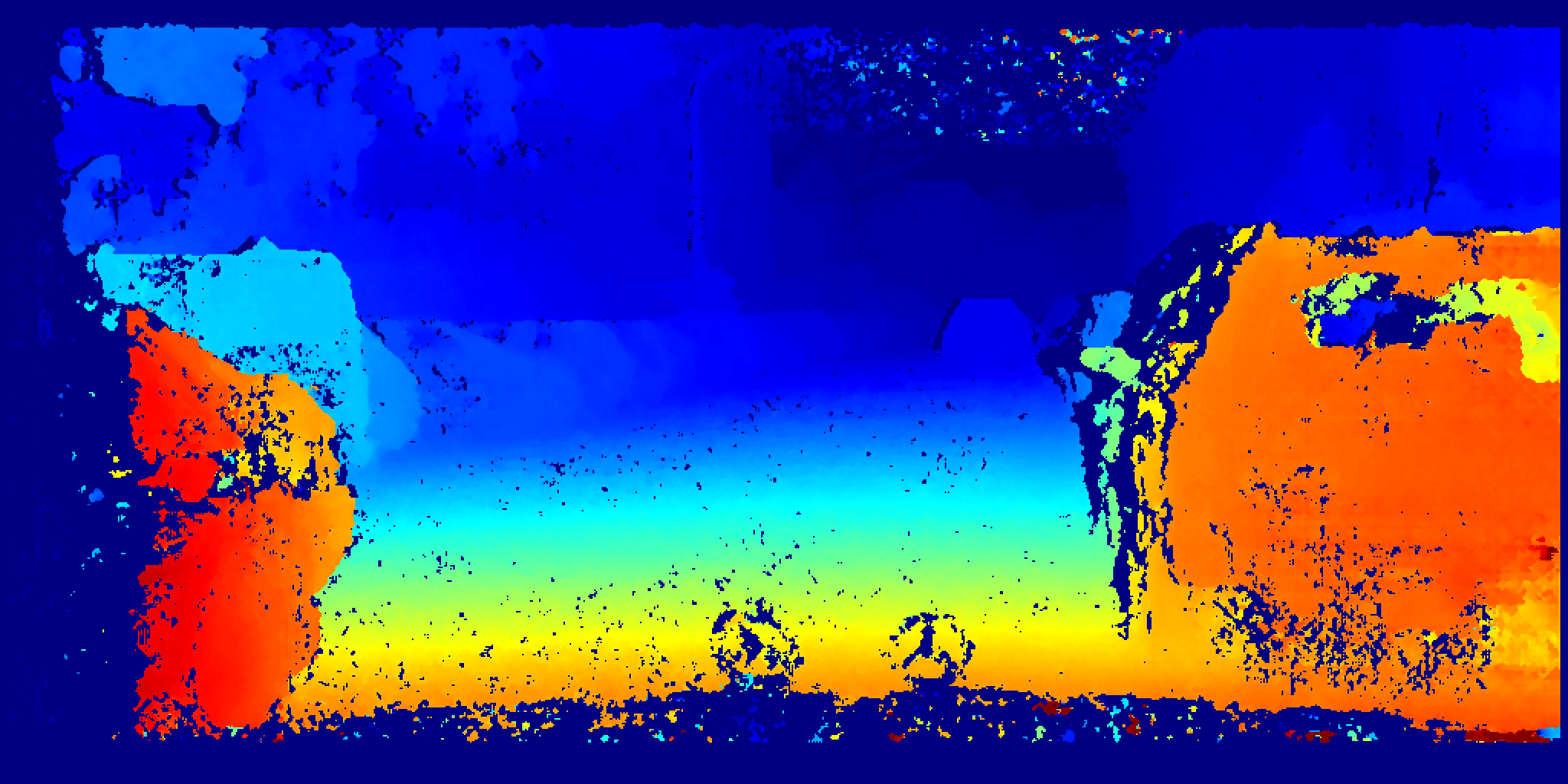}
    \caption{}
    \end{subfigure}%
    \begin{subfigure}{.5\textwidth}
    \centering
    \includegraphics[width=.95\linewidth]{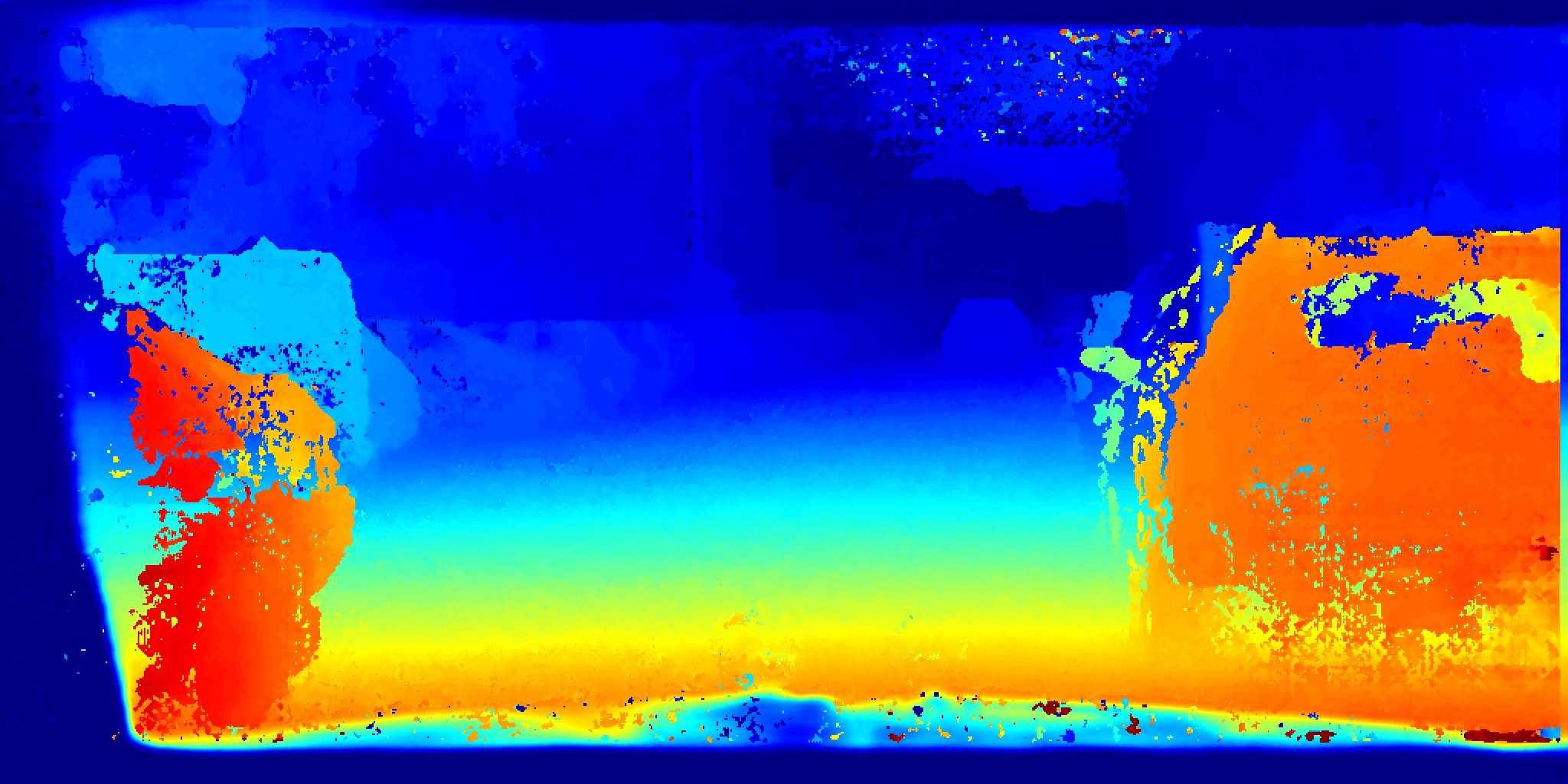}
    \caption{}
    \end{subfigure}

    \caption{ This figure illustrates the effect of the Iterative Learning Approach (ILA) on depth map refinement using samples from the Cityscapes dataset~\cite{Ref024,Ref025}. Figure (a) shows the original, unimproved depth map, which contains substantial missing and noisy regions resulting from sensor limitations. Figure(b) presents the corresponding depth map after applying the ILA, where missing depth values are iteratively filled using model predictions, leading to a more complete and structurally consistent depth representation. For visualization purposes, both depth maps are colorized~\cite{Ref026}, with warmer colors indicating closer regions and cooler colors representing farther distances.}
    \label{fig:fig01}
\end{figure*}

Through this iterative refinement, ILA improves depth estimation performance without requiring additional sensors or external annotations, relying instead on self-generated predictions to compensate for missing depth information (Figure \ref{fig:fig01}).

\subsection{Dataset}
\subsubsection{Cityscapes Dataset}
Cityscapes datasets are datasets captured from the European streets to collect RGB images, besides depth images and labels for the segmentation of images. 
We implemented the ILA Algorithm using these datasets, which is essential to generate a model for clear depth images without missing information (black pixels)
\subsubsection{Waymo Dataset}
The Waymo Open Dataset is a large-scale autonomous driving dataset collected using sensor-equipped Waymo vehicles operating in diverse urban and suburban environments. It provides high-resolution RGB images, LiDAR point clouds, and synchronized annotations for detection, tracking, and motion prediction tasks. The image quality and scene diversity make it suitable for robust perception research.

However, the dataset does not provide per-pixel dense depth maps aligned with the camera images. To address this limitation, we generated depth estimates using our in-house model developed with the ILA framework. The predicted depth maps were subsequently used in our pipeline to support downstream processing.

For experimental validation, we used 8,100 samples for training and 4,800 samples for validation. This split was designed to ensure sufficient data diversity while maintaining a reliable evaluation protocol.

\subsection{Metrics}
\subsection{Mean Average Precision at IoU = 0.50 (mAP@50)}
Mean Average Precision (mAP) is a standard evaluation metric for object detection tasks. It provides a comprehensive measure of detection performance by combining localization accuracy and classification correctness into a single scalar value. The metric is derived from the concept of Average Precision (AP), which is computed from the precision--recall curve for each object class. Precision represents the ratio of correctly predicted detections to the total number of predicted detections, while recall measures the ratio of correctly detected objects to the total number of ground-truth instances.

To determine whether a predicted bounding box is considered correct, the Intersection over Union (IoU) criterion is applied. IoU quantifies the overlap between a predicted bounding box and its corresponding ground-truth bounding box. It is defined as the ratio between the area of intersection and the area of union of the two bounding boxes. In the case of mAP@50, a detection is classified as a true positive if the IoU is greater than or equal to 0.50. This threshold reflects a moderate localization requirement and is commonly used in real-time detection benchmarks, as it balances strictness and robustness.

For each object class, a precision--recall curve is generated by varying the confidence threshold of the detector. The Average Precision (AP) is calculated as the area under this curve. The mean Average Precision (mAP) is then obtained by averaging the AP values across all object classes.

In this work, mAP@50 was used as the primary validation metric to evaluate model performance. It was selected because it provides a clear and interpretable measure of detection quality under degraded visibility conditions, such as fog, where precise bounding box localization can be challenging due to reduced contrast and visibility. Using mAP@50 enabled consistent comparison between the fog-intensity-specific models and allowed objective assessment of performance improvements during the validation phase.

\subsection{Bounding Box}

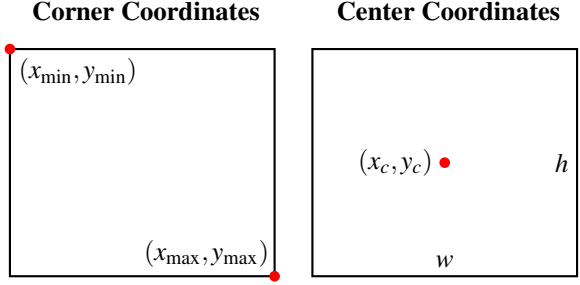
\begin{figure}[h]
\centering
\begin{tikzpicture}

% First rectangle (Corner format)
\draw[thick] (0,0) rectangle (3.5,3);
\fill[red] (0,3) circle (2pt);
\fill[red] (3.5,0) circle (2pt);

\node at (1.8,3.5) {
    
        \textbf{Corner Coordinates} 
};
\node at (0.9,2.7) {   
        $(x_{\min}, y_{\min})$ 
};
\node at (2.6,0.3) {
        $(x_{\max}, y_{\max})$
};
% Second rectangle (Center format)
\draw[thick] (4,0) rectangle (7.5,3);
\fill[red] (5.75,1.5) circle (2pt);
\node at (5.8,3.5) {
        \textbf{Center Coordinates} 
};
\node at (5.1,1.5) {      
        $(x_c, y_c)$ 
};
\node at (5.75,0.2) {
        $w$
};
\node at (7.3,1.5) {
        $h$
};
\end{tikzpicture}
\caption{Comparison between corner-coordinate and center-coordinate bounding box representations.}
\label{fig:fig02}
\end{figure}

A bounding box is a rectangular region used to localize and represent the spatial extent of an object within an image~\cite{Ref027}. In object detection tasks, bounding boxes are defined by their geometric parameters, typically either by the coordinates of the top-left and bottom-right corners $(x_{\min}, y_{\min}, x_{\max}, y_{\max})$, or by the center coordinates with width and height $(x_c, y_c, w, h)$ (Figure \ref{fig:fig02}). These parameters provide a compact numerical description of an object's position and scale in the image.

During training, ground-truth bounding boxes are annotated manually and serve as reference targets for the detection model. The model predicts bounding boxes along with class labels and confidence scores. The quality of a predicted bounding box is evaluated using the Intersection over Union (IoU) metric, which measures the overlap between the predicted box and the corresponding ground-truth box. A higher IoU indicates more accurate localization.
Bounding boxes play a fundamental role in object detection systems, as they directly influence both localization accuracy and evaluation metrics such as precision, recall, and mAP. In adverse weather conditions, including fog, accurate bounding box prediction becomes more challenging due to reduced visibility, contrast degradation, and blurred object boundaries. Therefore, improving bounding box regression under such conditions is essential for reliable perception in autonomous driving systems.

\subsection{Training Our Model}
For the purpose of this study, we extracted 8,100 images from the Waymo dataset for training and 4,800 images for validation. In addition, depth maps were generated using a model derived from the ILA algorithm. These depth maps are essential for the subsequent fog synthesis process.

The predicted depth representation follows the Cityscapes dataset encoding scheme. Specifically, disparity is computed as:

\begin{equation}\label{eq:02}
\text{disparity} = \frac{\text{float}(p) - 1}{256},
\end{equation}

where $p$ denotes the pixel value in the depth image, defined in the range $[0, 126]$. A value of $p = 0$ indicates an invalid measurement~\cite{Ref028}.

\begin{figure}[h]
    \centering
    \begin{subfigure}{.3\textwidth}
        \centering
        \includegraphics[width=.95\linewidth]{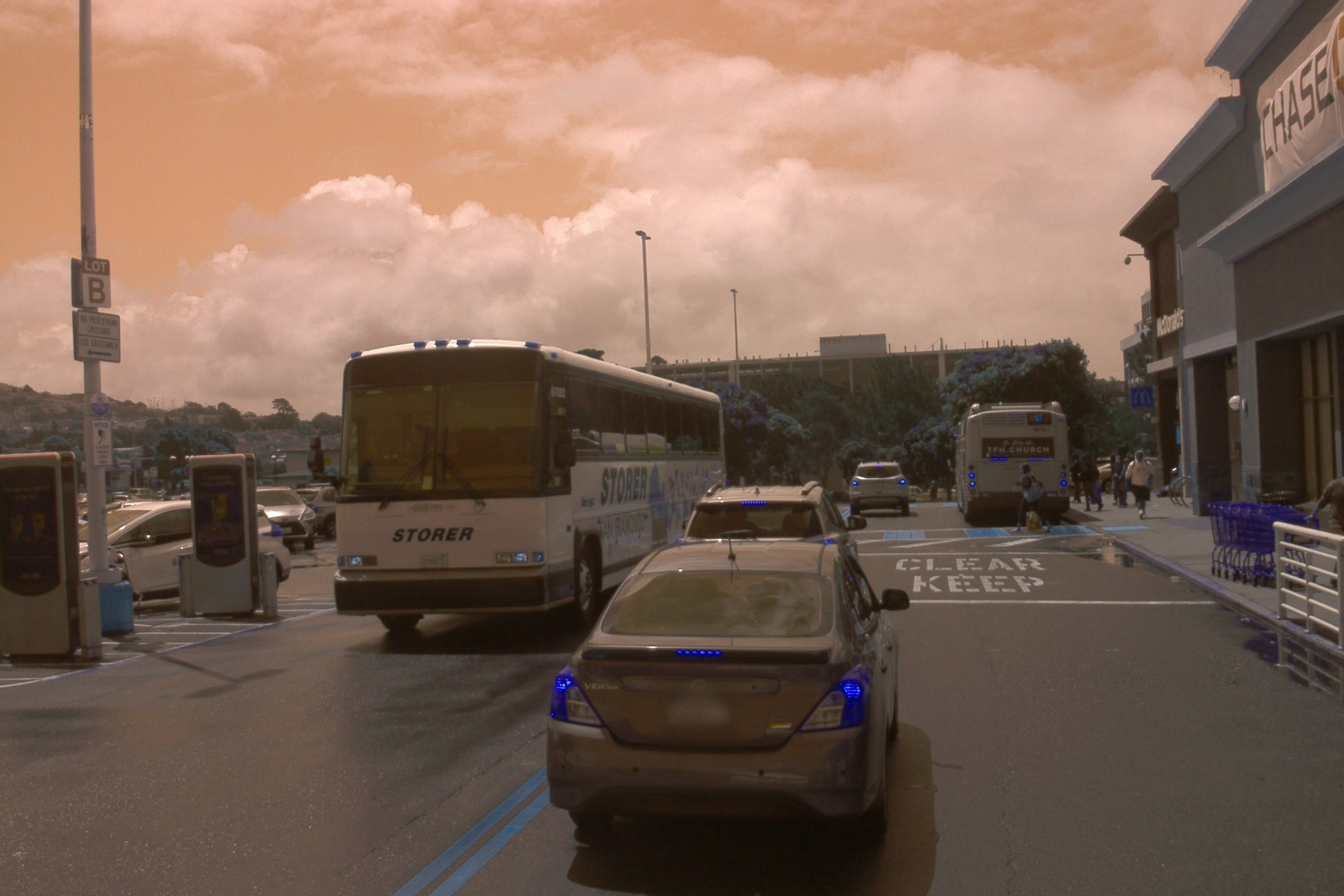}
        \caption{}
    \end{subfigure}%
    
    \begin{subfigure}{.3\textwidth}
        \centering
        \includegraphics[width=.95\linewidth]{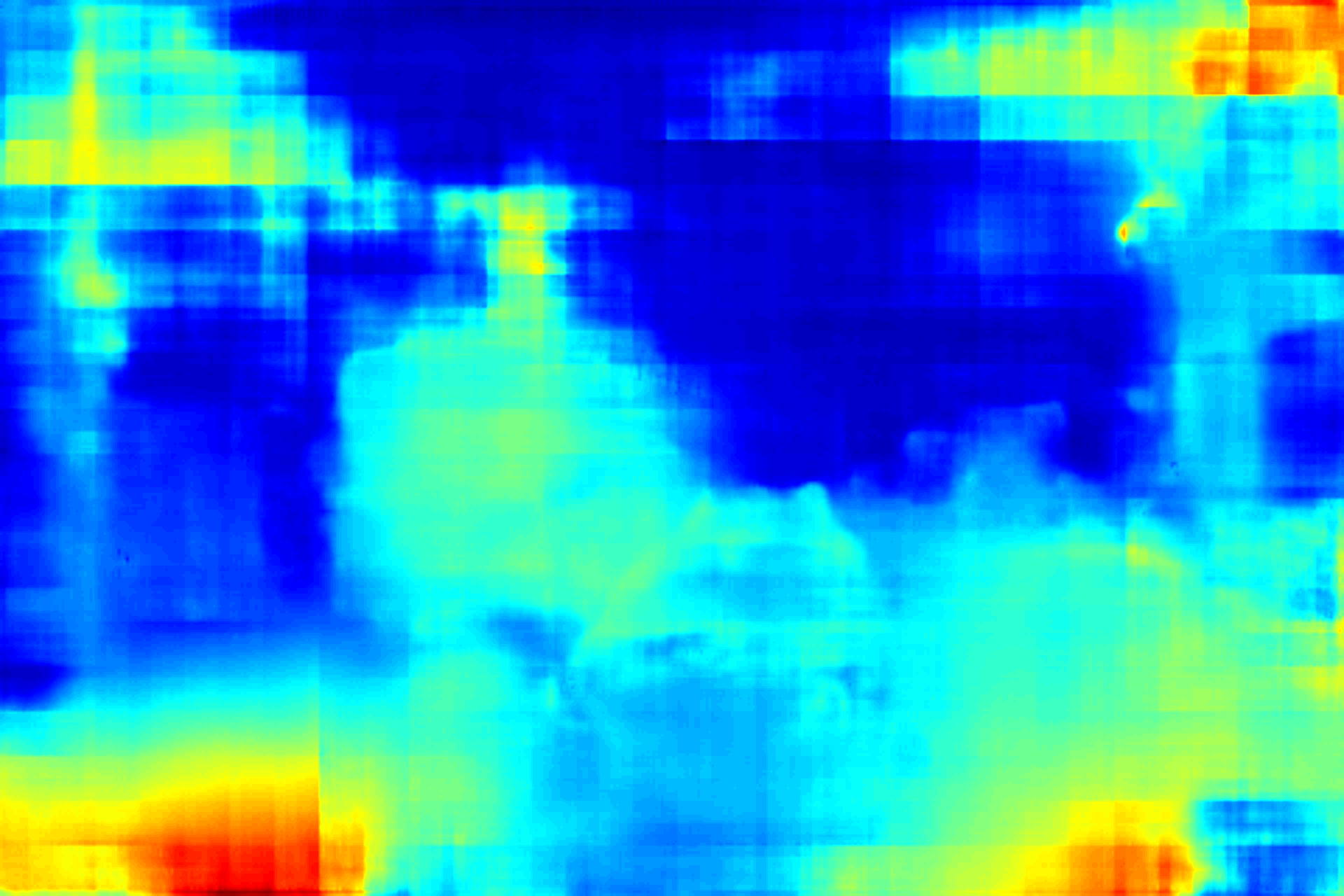}
        \caption{}
    \end{subfigure}
    
    \begin{subfigure}{.3\textwidth}
        \centering
        \includegraphics[width=.95\linewidth]{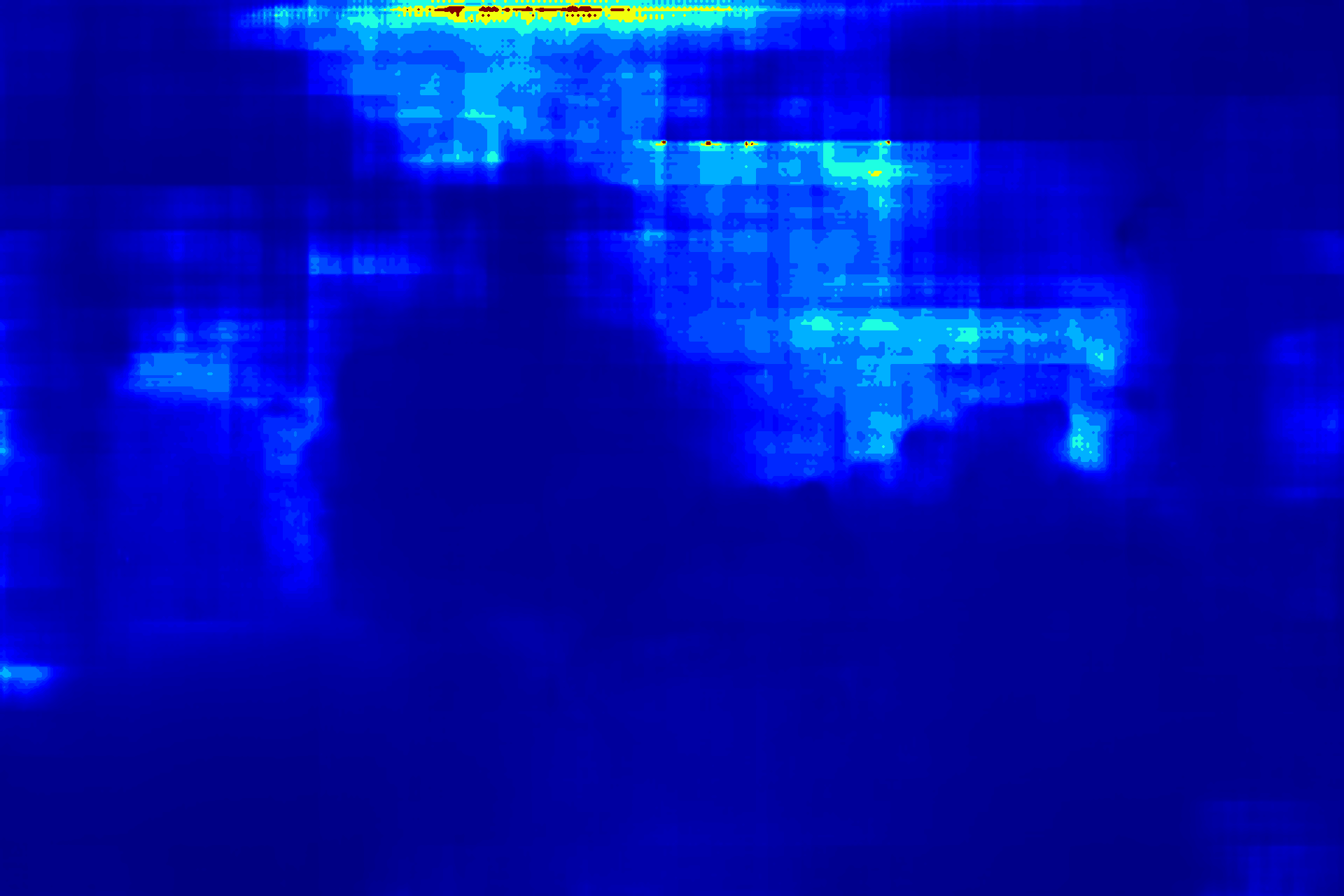}
        \caption{}
    \end{subfigure}%
    \caption{(a) Example RGB image from the Waymo dataset. (b) Colorized disparity map generated using the stereo camera configuration. (c) Colorized depth map after converting disparity values to metric distance (meters).}
    \label{fig:fig03}
\end{figure}

Using Eq.~\ref{eq:01}, and considering the stereo camera baseline of the Cityscapes dataset ($b = 22$ cm)~\cite{Ref024}, disparity values are converted into metric depth (in meters). This conversion enables the transformation of pixel-encoded disparity into physically meaningful distance measurements, as illustrated in Fig.~\ref{fig:fig03}.

In addition, the spatial resolution of the Waymo dataset images is $1920 \times 1280$ pixels.

\begin{figure*}[t]
\centering
\setlength{\tabcolsep}{2pt} 
\begin{tabular}{ccccc}
\includegraphics[width=0.19\linewidth]{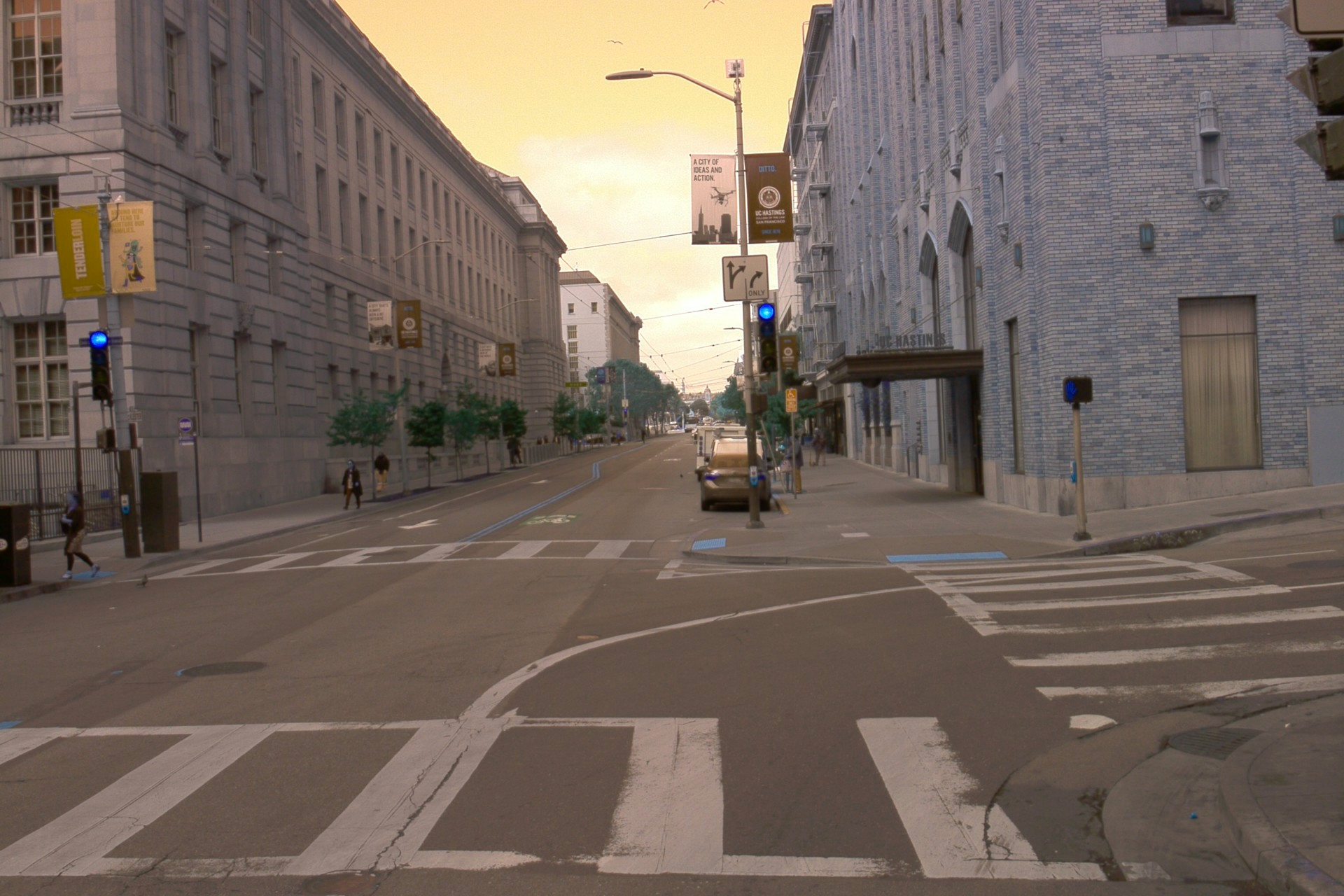} &
\includegraphics[width=0.19\linewidth]{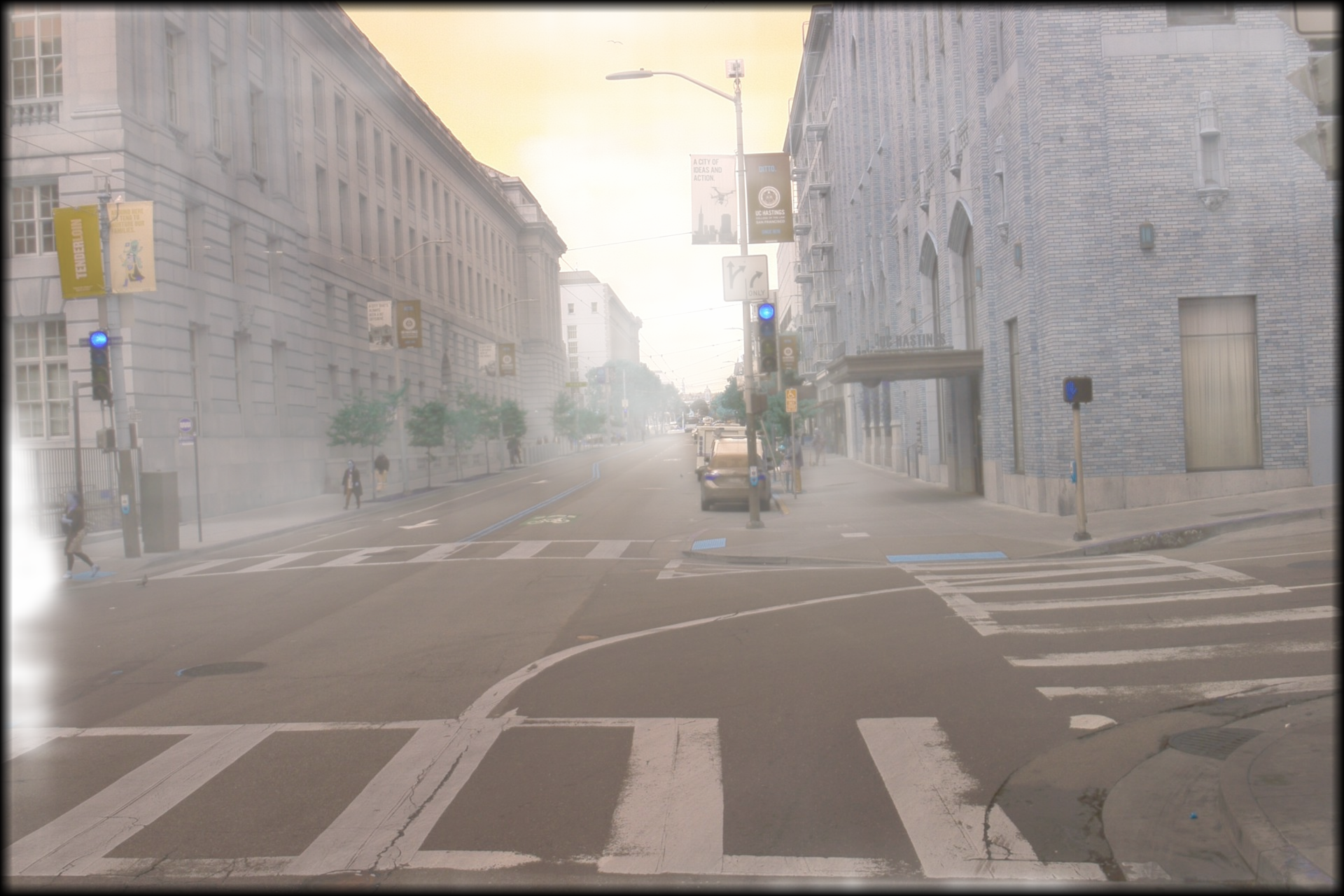} &
\includegraphics[width=0.19\linewidth]{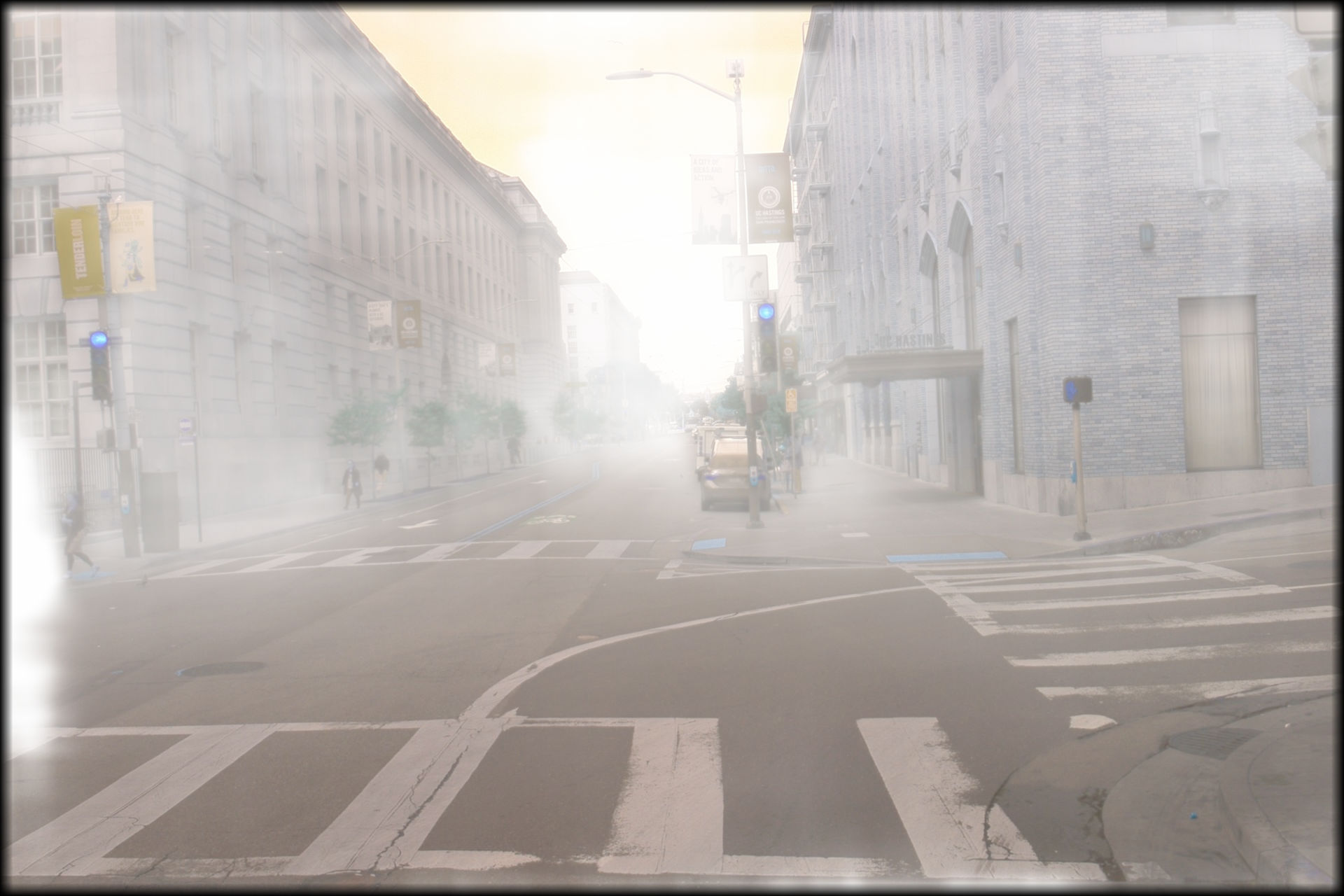} &
\includegraphics[width=0.19\linewidth]{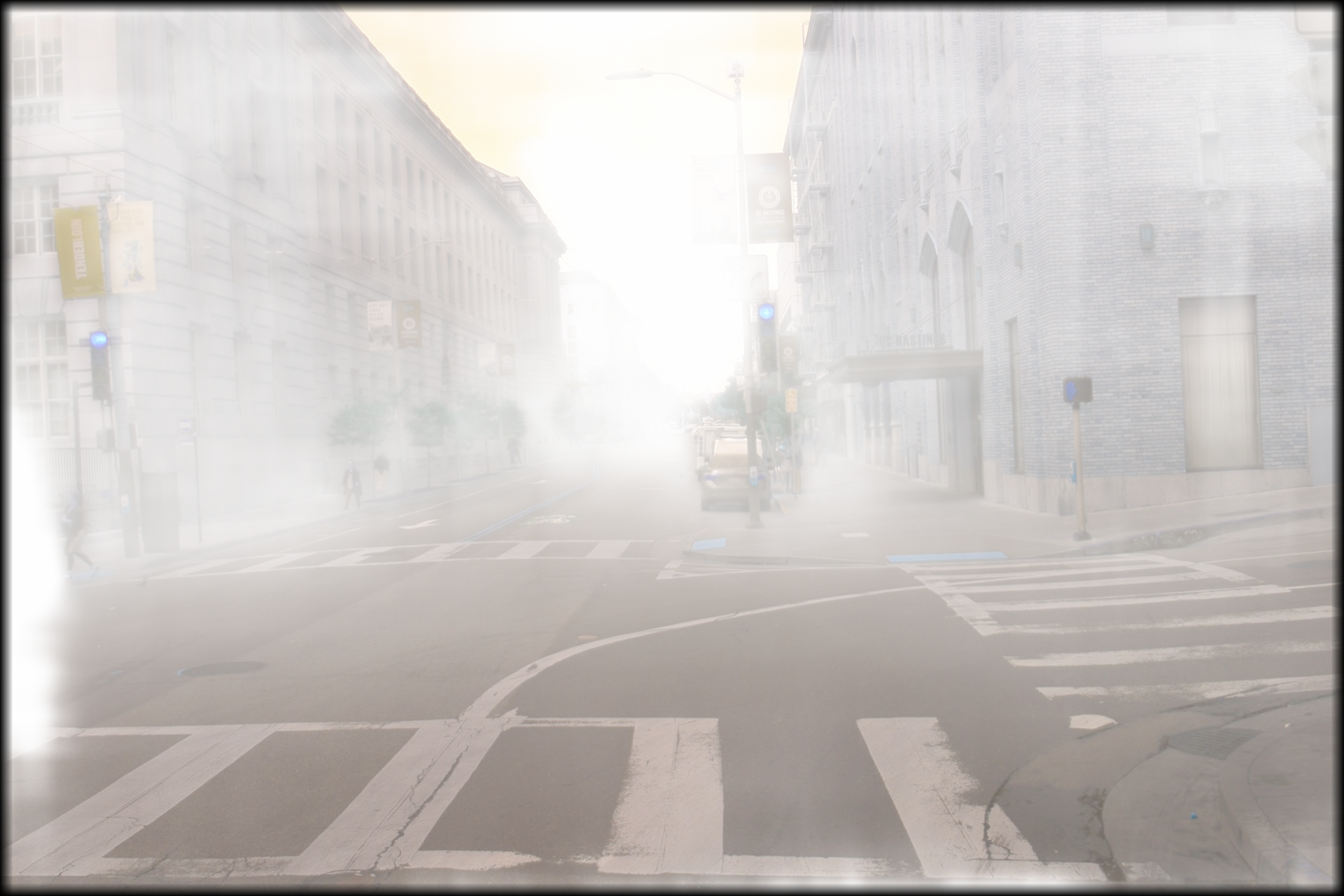} &
\includegraphics[width=0.19\linewidth]{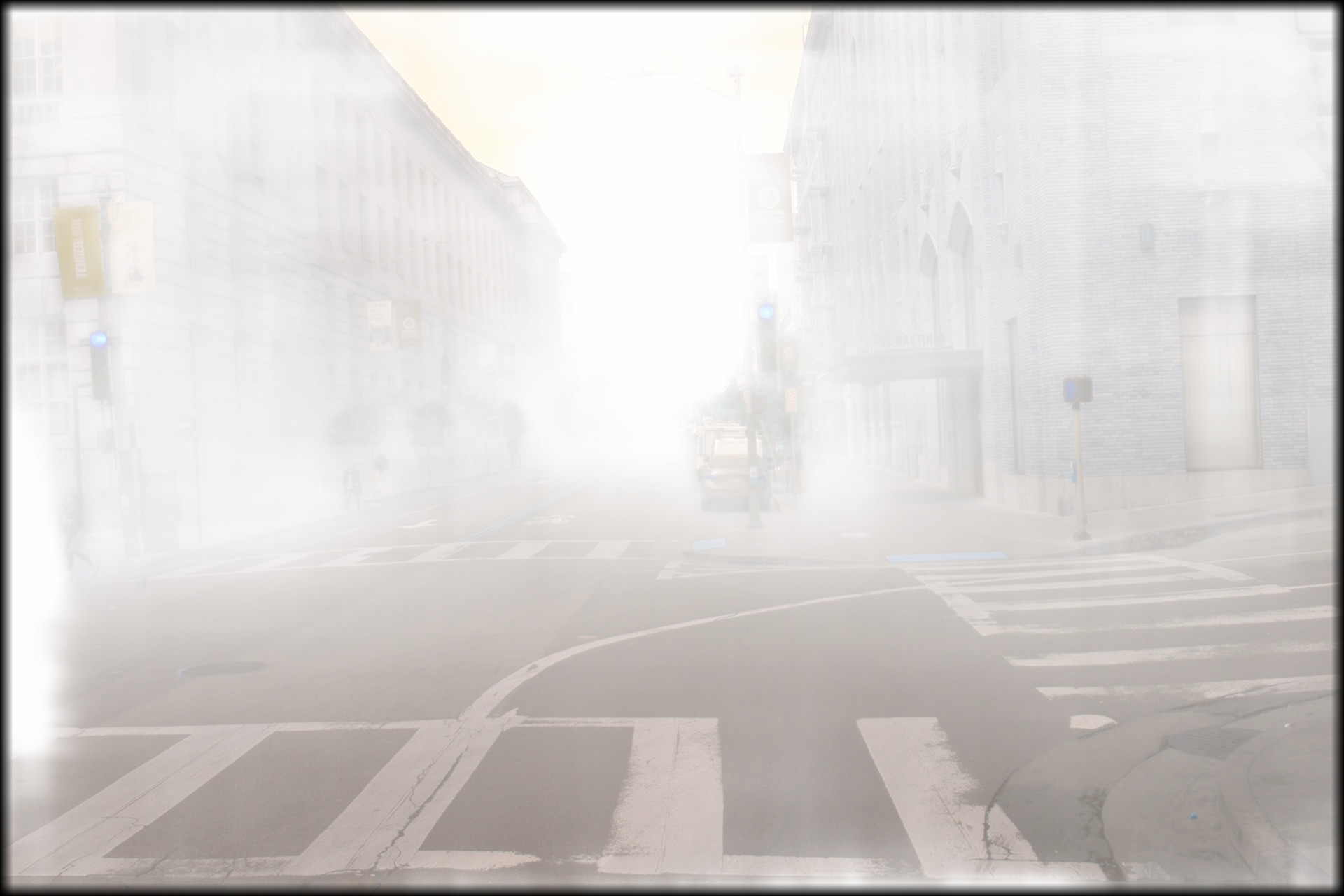} \\
{\footnotesize (a)} &
{\footnotesize (b) $\beta=0.03$} &
{\footnotesize (c) $\beta=0.06$} &
{\footnotesize (d) $\beta=0.09$} &
{\footnotesize (e) $\beta=0.12$}
\end{tabular}
\caption{Example image from the Waymo dataset. (a) Original clear-weather image. (b–e) Synthesized fog conditions generated using the extended Koschmieder formulation with $\beta = 0.03, 0.06, 0.09,$ and $0.12$, respectively.}
\label{fig:fig04}
\end{figure*}

To synthesize fog, we categorized atmospheric conditions according to the attenuation coefficient $\beta$ in the extended Koschmieder model. Specifically, we considered $\beta = 0.03$, $\beta = 0.06$, $\beta = 0.09$, and $\beta = 0.12$, corresponding to light fog, moderate fog, heavy fog, and very heavy fog, respectively (Fig.~\ref{fig:fig04}). The original clear-weather images were retained as an additional class. 

Consequently, five distinct weather conditions (clear, light fog, moderate fog, heavy fog, and very heavy fog) were defined as the basis for our experimental evaluation. To rigorously validate the proposed methodology, each weather condition was used to train a separate model. Performance was then evaluated using the corresponding test set for the same fog intensity level, allowing controlled assessment of perception performance under varying atmospheric degradations.

These images illustrate the four fog intensity levels considered in this study. For object detection, we employed the YOLOv12m model~\cite{Ref029}. Training was conducted for 100 epochs with an input image size of $640 \times 640$ pixels. The model was trained independently for each fog class, in addition to the clear-weather dataset, enabling a systematic evaluation of detection performance under progressively increasing fog density.

\subsection{Results and Discussions}

\begin{figure}[h]
    \centering
    \begin{subfigure}{.25\textwidth}
    \centering
    \includegraphics[width=.95\linewidth]{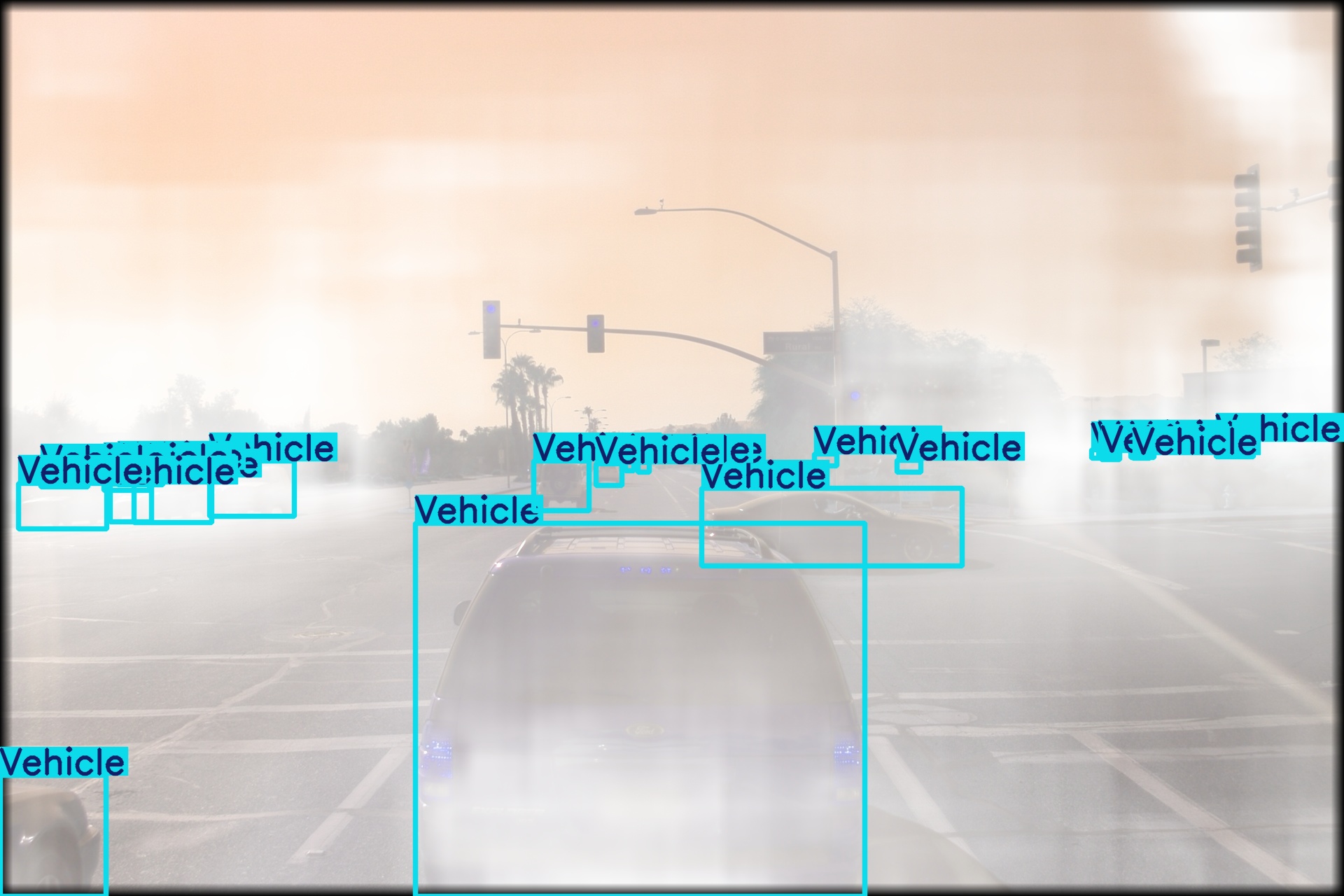} % RGB Image
    \caption{}
    \label{fig:sub10_1}
    \end{subfigure}%
    \begin{subfigure}{.25\textwidth}
    \centering
    \includegraphics[width=.95\linewidth]{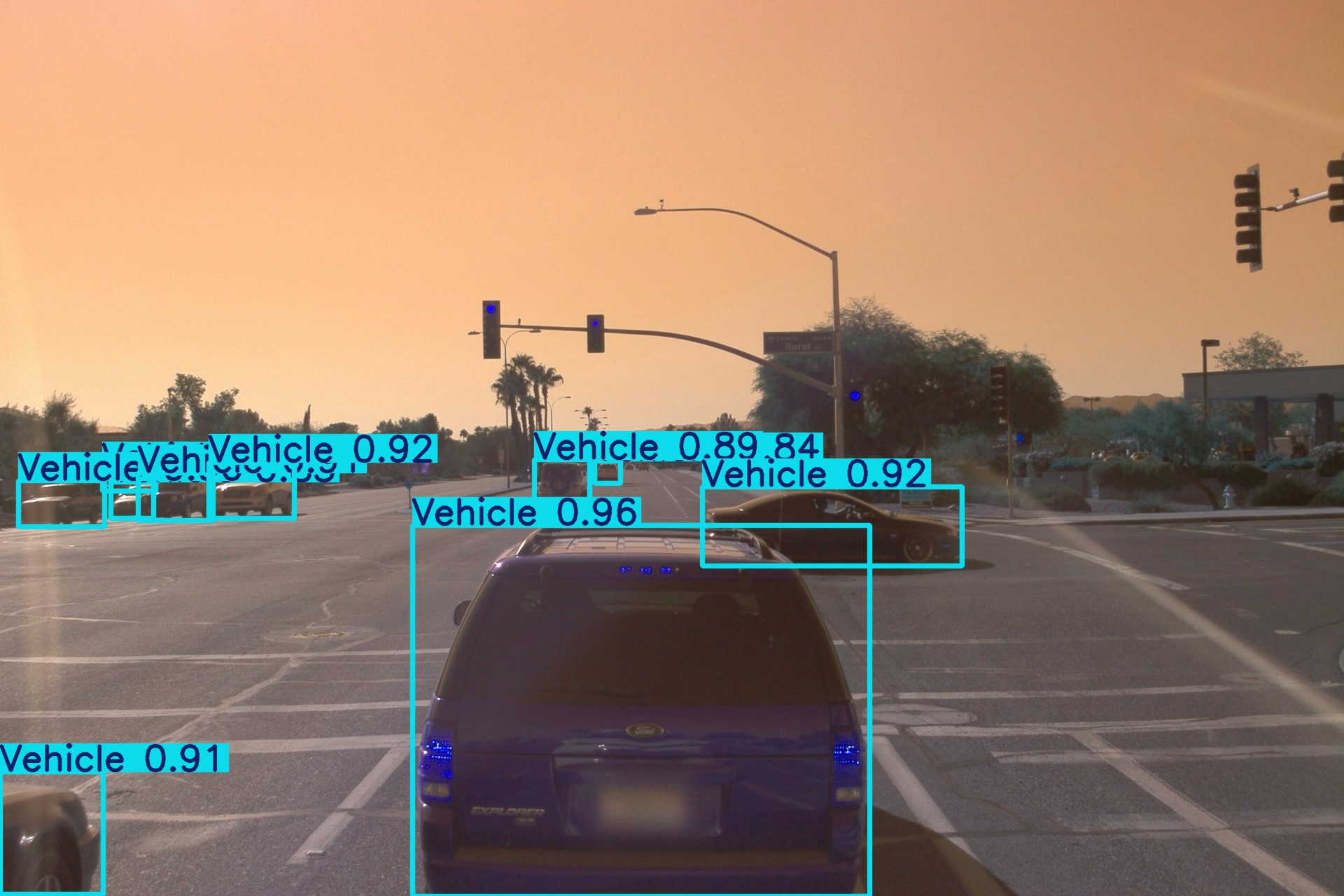} % Depth Image from cityscape
    \caption{}
    \label{fig:sub10_2}
    \end{subfigure}
    
    \begin{subfigure}{.25\textwidth}
    \centering
    \includegraphics[width=.95\linewidth]{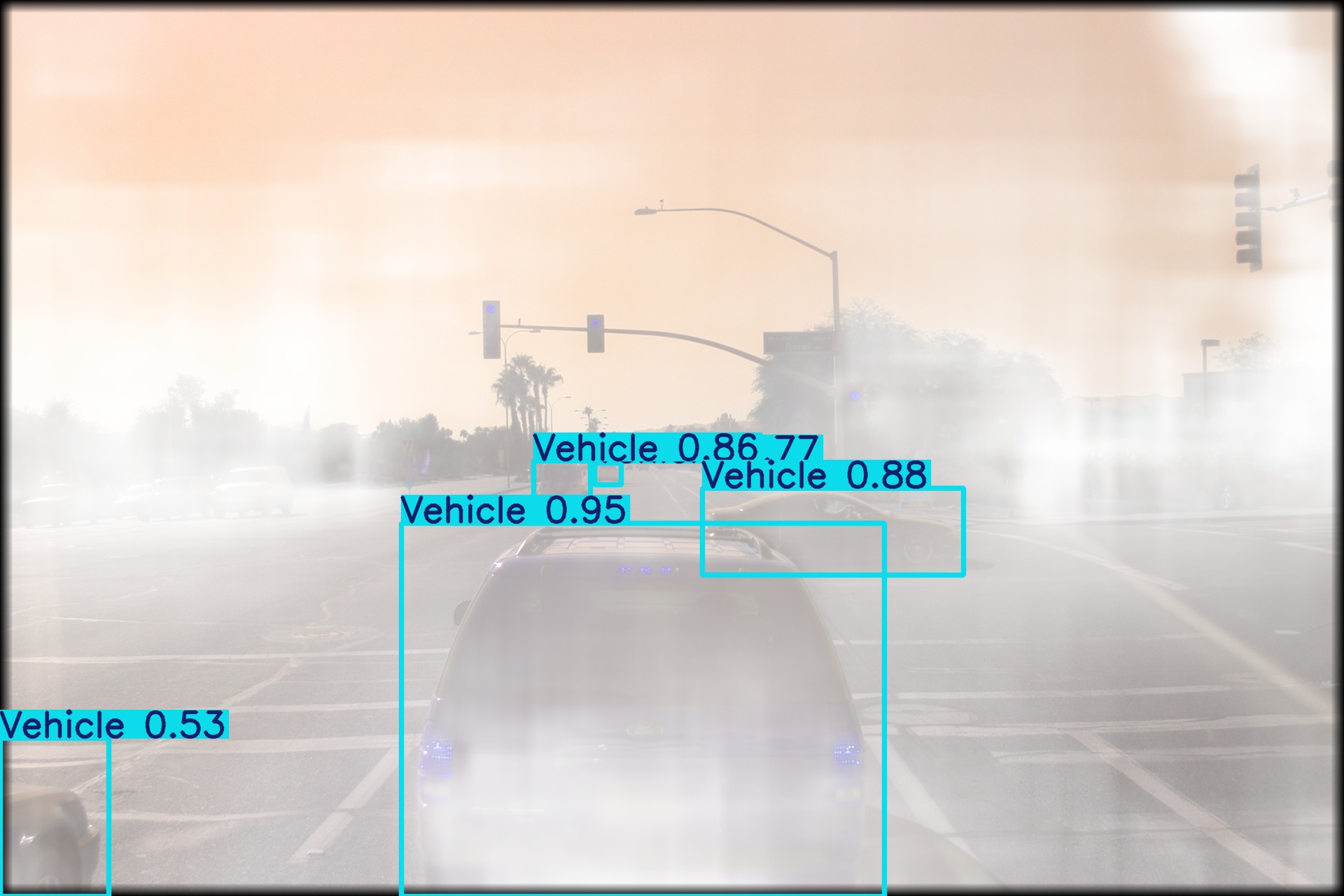} % Depth Image inhanced using AI
    \caption{}
    \label{fig:sub10_3}
    \end{subfigure}%
    \begin{subfigure}{.25\textwidth}
    \centering
    \includegraphics[width=.95\linewidth]{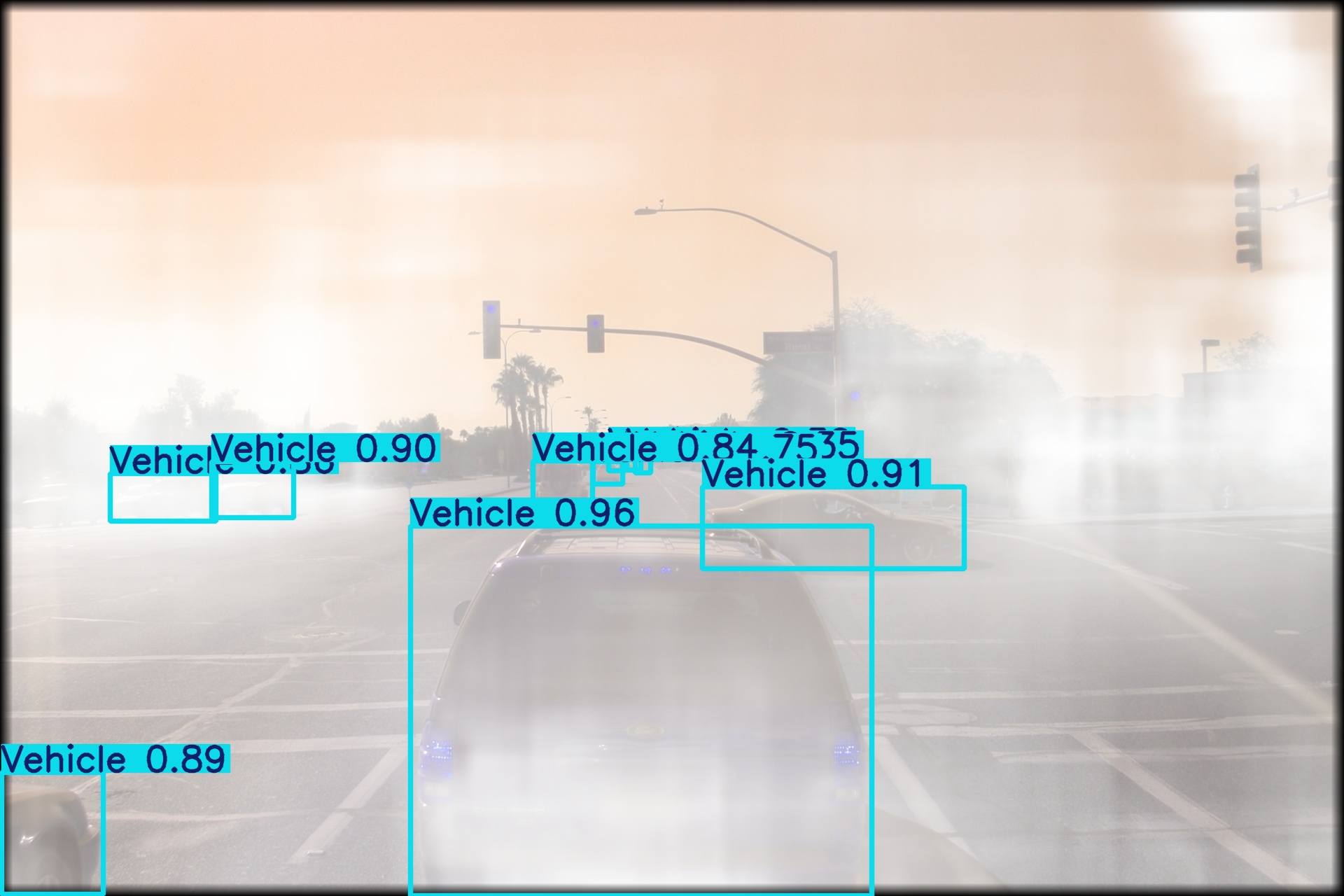} % Depth Image inhanced using loop on pixels
    \caption{}
    \label{fig:sub10_4}
    \end{subfigure}
    
    \caption{Qualitative comparison of detection results under clear and very heavy fog conditions ($\beta=0.12$). 
(a) Ground-truth bounding boxes for a sample from the Waymo dataset. 
(b) Predictions on the clear image using a model trained on clear-weather data. 
(c) Predictions on the same scene after synthesizing very heavy fog ($\beta=0.12$), using the model trained on clear-weather data. 
(d) Predictions on the foggy image using a model trained specifically for very heavy fog ($\beta=0.12$). 
The density-specific model demonstrates improved object retrieval under severe visibility degradation, with an increase in correctly detected objects compared to the clear-trained model.}
    \label{fig:fig05}
\end{figure}

Table~\ref{tab:tab2} presents a cross-density evaluation in which each model, trained on a specific fog level, is validated across all fog densities. The results indicate that the proposed density-aware training strategy yields particularly promising performance under light and moderate fog conditions ($\beta=0.03$ and $\beta=0.06$). In these scenarios, models trained on the corresponding fog levels achieve the highest or near-highest Recall and mAP@50 values, demonstrating that adapting the training distribution to the expected visibility level enhances detection robustness (Fig.~\ref{fig:fig05}).

As fog density increases, overall performance gradually declines due to stronger atmospheric attenuation. Interestingly, under very heavy fog conditions ($\beta=0.12$), the model trained on moderate fog ($\beta=0.06$) achieves competitive—and in some cases slightly higher—mAP@50 compared to the model trained specifically on very heavy fog. This observation suggests that extreme fog may introduce excessive visual degradation during training, potentially limiting the generalization capability of a model trained exclusively on the most severe condition.

From a practical deployment perspective, these findings indicate that a fully specialized model for each fog density may not be strictly necessary. Instead, a modular architecture composed of three models—one for clear weather, one for light fog, and one for moderate fog—may provide an effective balance between robustness and system complexity, while still maintaining competitive performance even under very heavy fog conditions.

Overall, the results confirm that fog density should be modeled explicitly, as performance varies systematically across density levels. While density-specific training provides measurable gains—especially in light and moderate fog—extreme fog conditions remain inherently challenging due to the substantial loss of visual information.

\begin{table}[t]
\caption{Cross-density evaluation of fog-specific models. Each row represents a model trained on a particular fog density ($\beta$), while each column corresponds to the validation dataset with a specific fog level. The highest value in each column is highlighted, indicating that performance generally peaks when the training and validation fog densities are matched or closely aligned. Evaluation is conducted using Recall and mAP@50 for object instances within 100\,m using YOLOv12m.}
\scriptsize
\begin{tabularx}{\linewidth}{cc|c|ccccc}
\toprule%
&\multirow{2}{3em}{\textbf{Metrics}\vspace{-3pt}}&&\multicolumn{5}{c}{\textbf{Validation Data}}\\
\cmidrule{3-8}
 &  & \textbf{$\beta$} &\textbf{ - }& \textbf{0.03 }&\textbf{ 0.06 }& \textbf{0.09 }& \textbf{0.12}\\
\midrule
\multirow{10}{1em}{\rotatebox[origin=c]{90}{\textbf{Trained model}}} &\multirow{5}{1em}{\rotatebox[origin=c]{90}{\textbf{Recall}}} & 
\textbf{ - } & \textbf{0.363} & 0.276 & 0.196 & 0.123 & 0.076 \\
&& \textbf{0.03 } & \textbf{0.346} & 0.337 & 0.304 & 0.265 & 0.225\\
&& \textbf{ 0.06 } & 0.307 & \textbf{0.32} & 0.302 & 0.279 & 0.242\\
&& \textbf{0.09 } & 0.283 & 0.301 & \textbf{0.392} & 0.275 & 0.254\\
&& \textbf{0.12} & 0.224 & 0.256 & \textbf{0.257} & 0.243 & 0.232\\
\cmidrule{2-8}
&\multirow{5}{1em}{\rotatebox[origin=c]{90}{\textbf{mAP50}}} & 
\textbf{ - } & \textbf{0.424} & 0.338 & 0.255 & 0.188 & 0.141 \\
&& \textbf{0.03 } & \textbf{0.407} & 0.397 & 0.358 & 0.311 & 0.267\\
&& \textbf{ 0.06 } & 0.377 & \textbf{0.381} & 0.358 & 0.327 & 0.293\\
&& \textbf{0.09 } & 0.336 & \textbf{0.357} & 0.347 & 0.324 & 0.295\\
&& \textbf{0.12} & 0.268 & \textbf{0.295} & 0.293 & 0.282 & 0.265\\
\bottomrule
\end{tabularx}\label{tab:tab2}
\end{table}

\section{\uppercase{Conclusions}}
\label{sec:conclusion}

In this work, we investigated a multi-density fog modeling strategy for object detection under adverse weather conditions. The experimental results demonstrate that density-aware training improves perception performance, particularly in terms of Recall, which is critical for detecting distant or partially occluded objects and reducing safety risks in foggy environments.

The findings further indicate that models trained for light and moderate fog conditions achieve consistently strong performance and, in some cases, provide competitive or even superior accuracy compared to models trained exclusively on very heavy fog. This suggests that extreme fog conditions introduce substantial visual degradation, which may limit the benefits of highly specialized training. From a practical deployment perspective, these results support a modular approach in which a limited set of models—such as clear-weather, light-fog, and moderate-fog models—can provide an effective balance between robustness and system complexity.

Although this study focused on RGB-based object detection using bounding boxes, the proposed density-aware modeling strategy can be extended to additional sensing modalities, including LiDAR and depth-based perception systems. Integrating multi-sensor information may further enhance robustness under severe atmospheric attenuation and represents a promising direction for future autonomous driving research.

\bibliographystyle{apalike}
{\small
\bibliography{bibliography}}

\begin{thebibliography}{}

\bibitem[Bijelic et~al., 2020]{Ref016}
Bijelic, M., Gruber, T., Mannan, F., Kraus, F., Ritter, W., Dietmayer, K., and Heide, F. (2020).
\newblock Seeing through fog without seeing fog: Deep multimodal sensor fusion in unseen adverse weather.
\newblock In {\em Proceedings of the IEEE/CVF Conference on Computer Vision and Pattern Recognition}, pages 11682--11692.

\bibitem[Chaar et~al., 2024]{Ref010}
Chaar, M.~M., Raiyn, J., and Weidl, G. (2024).
\newblock Improving the perception of objects under daylight foggy conditions in the surrounding environment.
\newblock {\em Vehicles}, 6(4):2154--2169.

\bibitem[Chaar et~al., 2025a]{Ref019}
Chaar, M.~M., Raiyn, J., and Weidl, G. (2025a).
\newblock Improve bounding box in carla simulator.
\newblock {\em arXiv preprint arXiv:2509.16773}.

\bibitem[Chaar et~al., 2025b]{Ref012}
Chaar, M.~M., Raiyn, J., and Weidl, G. (2025b).
\newblock Predicting depth maps from single rgb images and addressing missing information in depth estimation.
\newblock {\em arXiv preprint arXiv:2509.17686}.

\bibitem[Chaar et~al., 2023]{Ref003}
Chaar, M.~M., Weidl, G., and Raiyn, J. (2023).
\newblock Analyse the effect of fog on the perception.
\newblock {\em EU Science Hub: Brussels, Belgium}, page 332.

\bibitem[Cordts et~al., 2016]{Ref024}
Cordts, M., Omran, M., Ramos, S., Rehfeld, T., Enzweiler, M., Benenson, R., Franke, U., Roth, S., and Schiele, B. (2016).
\newblock The cityscapes dataset for semantic urban scene understanding.
\newblock In {\em Proceedings of the IEEE conference on computer vision and pattern recognition}, pages 3213--3223.

\bibitem[Cordts et~al., 2015]{Ref025}
Cordts, M., Omran, M., Ramos, S., Scharw{\"a}chter, T., Enzweiler, M., Benenson, R., Franke, U., Roth, S., and Schiele, B. (2015).
\newblock The cityscapes dataset.
\newblock In {\em CVPR Workshop on The Future of Datasets in Vision}.

\bibitem[Dosovitskiy et~al., 2017]{Ref020}
Dosovitskiy, A., Ros, G., Codevilla, F., Lopez, A., and Koltun, V. (2017).
\newblock Carla: An open urban driving simulator.
\newblock In Levine, S., Vanhoucke, V., and Goldberg, K., editors, {\em Proceedings of the 1st Annual Conference on Robot Learning}, volume~78 of {\em Proceedings of Machine Learning Research}, pages 1--16. PMLR.

\bibitem[Ganj et~al., 2025]{Ref013}
Ganj, A., Su, H., and Guo, T. (2025).
\newblock Hybriddepth: Robust metric depth fusion by leveraging depth from focus and single-image priors.
\newblock In {\em 2025 IEEE/CVF Winter Conference on Applications of Computer Vision (WACV)}, pages 973--982. IEEE.

\bibitem[Kaehler and Bradski, 2016]{Ref028}
Kaehler, A. and Bradski, G. (2016).
\newblock {\em Learning OpenCV 3: computer vision in C++ with the OpenCV library}.
\newblock " O'Reilly Media, Inc.".

\bibitem[Lai et~al., 2020]{Ref026}
Lai, C.-S., You, Z., Huang, C.-C., Tsai, Y.-H., and Chiu, W.-C. (2020).
\newblock Colorization of depth map via disentanglement.
\newblock In {\em Computer Vision--ECCV 2020: 16th European Conference, Glasgow, UK, August 23--28, 2020, Proceedings, Part VII 16}, pages 450--466. Springer.

\bibitem[Lee and Kim, 2017]{Ref008}
Lee, Y. and Kim, G. (2017).
\newblock Fog level estimation using non-parametric intensity curves in road environments.
\newblock {\em Electronics Letters}, 53(21):1404--1406.

\bibitem[Liu et~al., 2026]{Ref018}
Liu, Z., Zhang, J., Zhang, X., and Song, H. (2026).
\newblock Robust object detection in adverse weather conditions: Ecl-yolov11 for automotive vision systems.
\newblock {\em Sensors}, 26(1):304.

\bibitem[Mirza et~al., 2021]{Ref002}
Mirza, M.~J., Buerkle, C., Jarquin, J., Opitz, M., Oboril, F., Scholl, K.-U., and Bischof, H. (2021).
\newblock Robustness of object detectors in degrading weather conditions.
\newblock In {\em 2021 IEEE International Intelligent Transportation Systems Conference (ITSC)}, pages 2719--2724. IEEE.

\bibitem[Mohan et~al., 2026]{Ref014}
Mohan, A., Meena, H.~K., Wajid, M., and Srivastava, A. (2026).
\newblock Graph signal processing-based road object detection using mmwave radar for adas application.
\newblock {\em IEEE Internet of Things Journal}.

\bibitem[Niranjan et~al., 2021]{Ref021}
Niranjan, D., VinayKarthik, B., et~al. (2021).
\newblock Deep learning based object detection model for autonomous driving research using carla simulator.
\newblock In {\em 2021 2nd international conference on smart electronics and communication (ICOSEC)}, pages 1251--1258. IEEE.

\bibitem[Oeljeklaus, 2021]{Ref009}
Oeljeklaus, M. (2021).
\newblock {\em An integrated approach for traffic scene understanding from monocular cameras}.
\newblock VDI Verlag D{\"u}sseldorf, Germany.

\bibitem[Shen et~al., 2026]{Ref022}
Shen, Z., Lin, C., Nie, L., Liao, K., Lin, W., and Zhao, Y. (2026).
\newblock Revisiting 360 depth estimation with panogabor: A new fusion perspective.
\newblock {\em IEEE Transactions on Pattern Analysis and Machine Intelligence}.

\bibitem[Soumya et~al., 2026]{Ref011}
Soumya, A., Dubey, A.~K., Mohan, C.~K., and Cenkeramaddi, L.~R. (2026).
\newblock Point cloud-based multi-target 3d object detection using lidar sensor and deep learning.
\newblock {\em Journal of Ambient Intelligence and Humanized Computing}, pages 1--16.

\bibitem[Sun et~al., 2020]{Ref004}
Sun, P., Kretzschmar, H., Dotiwalla, X., Chouard, A., Patnaik, V., Tsui, P., Guo, J., Zhou, Y., Chai, Y., Caine, B., Vasudevan, V., Han, W., Ngiam, J., Zhao, H., Timofeev, A., Ettinger, S., Krivokon, M., Gao, A., Joshi, A., Zhang, Y., Shlens, J., Chen, Z., and Anguelov, D. (2020).
\newblock Scalability in perception for autonomous driving: Waymo open dataset.

\bibitem[Tian et~al., 2025]{Ref029}
Tian, Y., Ye, Q., and Doermann, D. (2025).
\newblock Yolov12: Attention-centric real-time object detectors.
\newblock {\em arXiv preprint arXiv:2502.12524}.

\bibitem[Valanarasu et~al., 2022]{Ref017}
Valanarasu, J. M.~J., Yasarla, R., and Patel, V.~M. (2022).
\newblock Transweather: Transformer-based restoration of images degraded by adverse weather conditions.
\newblock In {\em Proceedings of the IEEE/CVF Conference on Computer Vision and Pattern Recognition}, pages 2353--2363.

\bibitem[Vargas et~al., 2021]{Ref015}
Vargas, J., Alsweiss, S., Toker, O., Razdan, R., and Santos, J. (2021).
\newblock An overview of autonomous vehicles sensors and their vulnerability to weather conditions.
\newblock {\em Sensors}, 21(16):5397.

\bibitem[Xie et~al., 2024]{Ref007}
Xie, Y., Wei, H., Liu, Z., Wang, X., and Ji, X. (2024).
\newblock Synfog: a photo-realistic synthetic fog dataset based on end-to-end imaging simulation for advancing real-world defogging in autonomous driving.
\newblock In {\em Proceedings of the IEEE/CVF Conference on Computer Vision and Pattern Recognition}, pages 21763--21772.

\bibitem[Yao et~al., 2022]{Ref027}
Yao, Y., Cheng, G., Wang, G., Li, S., Zhou, P., Xie, X., and Han, J. (2022).
\newblock On improving bounding box representations for oriented object detection.
\newblock {\em IEEE Transactions on Geoscience and Remote Sensing}, 61:1--11.

\bibitem[Yurtsever et~al., 2020]{Ref001}
Yurtsever, E., Lambert, J., Carballo, A., and Takeda, K. (2020).
\newblock A survey of autonomous driving: Common practices and emerging technologies.
\newblock {\em IEEE access}, 8:58443--58469.

\bibitem[Zhao et~al., 2019]{Ref006}
Zhao, Z.-Q., Zheng, P., Xu, S.-t., and Wu, X. (2019).
\newblock Object detection with deep learning: A review.
\newblock {\em IEEE transactions on neural networks and learning systems}, 30(11):3212--3232.

\bibitem[Zilong et~al., 2026]{Ref023}
Zilong, X., Zhang, C., Dibin, W., Yan, X., and Zhao, Y. (2026).
\newblock Dynamic visibility recognition and driving risk assessment under rain--fog conditions using monocular surveillance imagery.
\newblock {\em Sustainability}, 18(2):625.

\bibitem[Zou et~al., 2023]{Ref005}
Zou, Z., Chen, K., Shi, Z., Guo, Y., and Ye, J. (2023).
\newblock Object detection in 20 years: A survey.
\newblock {\em Proceedings of the IEEE}, 111(3):257--276.

\end{thebibliography}

\end{document}